\documentclass[12pt, final]{l4dc2021}

\title[Benchmarking Energy-Conserving Neural Networks]{Benchmarking Energy-Conserving Neural Networks for Learning Dynamics from Data}
\usepackage{times}

\author{%
\Name{Yaofeng Desmond Zhong} 
\Email{yaofeng.zhong@siemens.com}
\\
\Name{Biswadip Dey} \Email{biswadip.dey@siemens.com}\\
 \Name{Amit Chakraborty} \Email{amit.chakraborty@siemens.com}\\
 \addr Siemens Corporation, 755 College Road East, Princeton, NJ 08536}

\usepackage{wrapfig}

\usepackage{soul}

\begin{document}

\maketitle

\begin{abstract}%
    The last few years have witnessed an increased interest in incorporating physics-informed inductive bias in deep learning frameworks. In particular, a growing volume of literature has been exploring ways to enforce energy conservation while using neural networks for learning dynamics from observed time-series data. In this work, we survey ten recently proposed energy-conserving neural network models, including HNN, LNN, DeLaN, SymODEN, CHNN, CLNN and their variants. We provide a compact derivation of the theory behind these models and explain their similarities and differences. Their performance are compared in 4 physical systems. We point out the possibility of leveraging some of these energy-conserving models to design energy-based controllers. 
\end{abstract}


%
%
%
\section{Introduction}
The motion of physical systems is governed by the laws of physics, which can be expressed as a set of  differential equations. These equations describe how the state of the system, e.g., the position and velocity of objects, evolve over time. Given such a set of differential equations and an initial state, we can solve the initial value problem for future states and predict the trajectory of the physical system. On the other hand, if the underlying dynamics, i.e., the differential equations, are unknown and a set of trajectory data is given instead, we can formulate a learning problem to infer the unknown dynamics that can explain the trajectory data and leverage the learned differential equations for making predictions based on a set of new initial states. From a learning perspective, deep neural networks are very effective function approximators \citep{pinkus1999approximation, liang2016deep} and recent advances in end-to-end representation learning have enabled their wide adoption in various application domains \citep{goodfellow2016deep}. To avoid overfitting and enable generalization beyond training data, the neural networks employ appropriate inductive bias by preferring certain representations over others \citep{haussler1988quantifying}; often, the computation graph is constructed to capture/reflect some salient aspects of the problem under consideration.

In recent years, a growing number of neural network models have been proposed to solve the problem of learning dynamics, expressed via a set of differential equations, from data \citep{lutter2018deep, greydanus2019hamiltonian, Zhong2020Symplectic, Chen2020Symplectic, roehrl2020modeling, cranmer2020lagrangian, finzi2020simplifying}. Most of these models are based on Neural ODE \citep{chen2018neural}, a neural network approach to learning an unknown set of differential equations from trajectory data. When it comes to learning the motion of physical systems, we can leverage the laws of physics for data efficiency and generalization. Two popular choices to model the relevant physics are the Lagrangian and Hamiltonian dynamics, which are both reformulations of Newton's second law and can model a broad class of physical systems. Moreover, both of these formalisms conserve energy under proper assumptions, which serves as a great physics-based prior when modeling energy-conserved system. In addition, there is a growing volume of work that focuses on enforcing physics while learning the underlying dynamics from high-dimensional image sequence or video data \citep{zhong2020unsupervised, allen2020lagnetvip}.

In this work, we benchmark ten recently introduced neural network models that impose structures on the underlying computation graph for learning Lagrangian or Hamiltonian dynamics from trajectory data. To aid informed selection of neural network model(s) for a given task, we focus on highlighting and explaining the similarities and differences between these models in 4 physical systems.
%
%
%

%
%
%
\section{Preliminaries}
%
Lagrangian and Hamiltonian dynamics are both reformulations of Newton's second law of motion. If holonomic constraints exist, those constraints can be enforced either implicitly or explicitly. The former is simpler to derive, while the latter simplifies the functional form. In this section, we review Lagrangian and Hamiltonian dynamics with both implicit and explicit constraints, derived from D'Alembert's principle (the principle of virtual work). Please see \cite{goldstein2002classical} for more details on the implicit constraint derivation, and \cite{date2010lectures}, \cite{finzi2020simplifying} for the explicit constraint derivation using variational principle. We wrap up this section by reviewing how to learn these dynamics from data.
%
%

%
%
\subsection{Lagrangian dynamics with implicit constraints}
%
For a rigid body system with degrees of freedom (DOF) $M$, the configuration of rigid bodies at time $t$ can be described by independent generalized coordinates $\mathbf{q}(t)=(q_1(t), q_2(t), ..., q_M(t))$. From D'Alembert's principle (the principle of virtual work), Newton's second law can be expressed as 
\begin{equation}
\label{eq:D}
    \delta \mathbf{q}^{T} \Big[\frac{\mathrm{d}}{\mathrm{d}t} \Big( \frac{\partial L}{\partial \dot{\mathbf{q}}} \Big) - \frac{\partial L}{\partial \mathbf{q}}\Big] = 0,
\end{equation}
where the scalar function $L(\mathbf{q}, \dot{\mathbf{q}})$ is the Lagrangian, $\dot{\mathbf{q}} = \mathrm{d}\mathbf{q}/\mathrm{d}t$ and $\delta \mathbf{q}$ is the virtual displacement. Here we assume all the generalized forces are conservative which can be captured by a potential field. 
As the $M$ generalized coordinates are independent, each $\delta q_i$ can be varied independently. Then the only way for Equation \eqref{eq:D} to hold is the following set of $M$ equations, 
\begin{equation}
\label{eq:EL}
    \frac{\mathrm{d}}{\mathrm{d}t} \Big( \frac{\partial L}{\partial \dot{\mathbf{q}}} \Big) - \frac{\partial L}{\partial \mathbf{q}} = \mathbf{0},
\end{equation}
which is also referred to as the Euler-Lagrange equation.
We can symbolically expand the Euler-Lagrange equation into a set of $M$ second-order differential equations,
\begin{equation}
\label{eq:2nd-order-lag}
\ddot{\mathbf{q}} = (\nabla_{\dot{\mathbf{q}}}\nabla_{\dot{\mathbf{q}}}^T L )^{-1} [\nabla_{\mathbf{q}} L - (\nabla_{\mathbf{q}}\nabla_{\dot{\mathbf{q}}}^T L)\dot{\mathbf{q}}],
\end{equation}
where the operators are defined as $(\nabla_{\dot{\mathbf{q}}})_i = \frac{\partial}{\partial \dot{q}_i}$, $ (\nabla_{\mathbf{q}}\nabla_{\dot{\mathbf{q}}}^T)_{ij} = \frac{\partial^2}{\partial q_j \partial \dot{q}_i}$ and $(\nabla_{\dot{\mathbf{q}}}\nabla_{\dot{\mathbf{q}}}^T)_{ij} = \frac{\partial^2}{\partial \dot{q}_j \partial \dot{q}_i}$.

The physical meaning of the Lagrangian $L(\mathbf{q}, \dot{\mathbf{q}})$ is the difference between kinetic energy $T(\mathbf{q}, \dot{\mathbf{q}})$ and potential energy $V(\mathbf{q})$. For rigid body systems, the Lagrangian is 
\begin{equation}
\label{eq:Lag}
    L(\mathbf{q}, \dot{\mathbf{q}}) = T(\mathbf{q}, \dot{\mathbf{q}}) - V(\mathbf{q}) = \frac{1}{2} \dot{\mathbf{q}}^T \mathbf{M}(\mathbf{q})\dot{\mathbf{q}} - V(\mathbf{q}),
\end{equation}
where $\mathbf{M}(\mathbf{q})$ is the positive definite mass matrix. With this form, Equation \eqref{eq:2nd-order-lag} can be written as 
\begin{equation}
\label{eq:2nd-order-lag-s}
\ddot{\mathbf{q}} = \mathbf{M}(\mathbf{q})^{-1} [- \dot{\mathbf{M}}(\mathbf{q}) \dot{\mathbf{q}} +\frac{1}{2} \nabla_\mathbf{q} \Big( \dot{\mathbf{q}}^T \mathbf{M}(\mathbf{q})\dot{\mathbf{q}} \Big) - \nabla_\mathbf{q} V(\mathbf{q})].
\end{equation}
%
%

%
%
\subsection{Hamiltonian dynamics with implicit constraints}
%
The Hamiltonian dynamics can be derived by performing Legendre transformation on the Lagrangian to get the conjugate momentum $\mathbf{p} = \nabla_{\dot{\mathbf{q}}} L$ and the Hamiltonian $H(\mathbf{q}, \mathbf{p}) = \dot{\mathbf{q}}^T \mathbf{p} - L(\mathbf{q}, \dot{\mathbf{q}})$. The Euler-Lagrangian equation \eqref{eq:EL} can then be expressed as the following Hamiltonian dynamics
\begin{equation}
    \dot{\mathbf{q}} = \nabla_{\mathbf{p}} H, \qquad 
    \dot{\mathbf{p}} = -\nabla_{\mathbf{q}} H.
    \label{eq:ham}
\end{equation}
Along a trajectory of Hamiltonian dynamics, the Hamiltonian $H$ is conserved, since
\begin{equation}
    \label{eq:H_conserve}
    \dot{H} = (\nabla_{\mathbf{q}}^T H)  \dot{\mathbf{q}} + (\nabla_{\mathbf{p}}^T H) \dot{\mathbf{p}} = 0.
\end{equation}

For a rigid body system, we can use the structured Lagrangian in Equation \eqref{eq:Lag} and get the conjugate momentum $\mathbf{p} = \mathbf{M}(\mathbf{q}) \dot{\mathbf{q}}$ and the structured Hamiltonian 
\begin{equation}
H(\mathbf{q}, \mathbf{p}) = \frac{1}{2}\mathbf{p}^T \mathbf{M}^{-1}(\mathbf{q}) \mathbf{p} + V(\mathbf{q}),
\label{eqn:Ham}
\end{equation}
The Hamiltonian equals the total energy of the rigid body system and Equation \eqref{eq:H_conserve} shows the energy conservation property of Hamiltonian dynamics. With the structured Hamiltonian \eqref{eqn:Ham}, we can express the Hamiltonian dynamics as 
\begin{equation}
    \dot{\mathbf{q}} = \mathbf{M}^{-1}(\mathbf{q}) \mathbf{p} , \qquad 
    \dot{\mathbf{p}} = -\frac{1}{2}\nabla_\mathbf{q}  \Big(\mathbf{p}^T \mathbf{M}^{-1}(\mathbf{q}) \mathbf{p}\Big) - 
    \nabla_\mathbf{q} V(\mathbf{q})
    \label{eq:1st-order-ham-s}
\end{equation}
%
%

%
%
\subsection{Lagrangian dynamics with explicit constraints}
%
The derivations above enforce holonomic constraints implicitly by choosing a number of generalized coordinates that equals the degrees of freedom $M$. To derive Lagrangian and Hamiltonian dynamics, we can also enforce the holonomic constraints explicitly so that we can use Cartesian coordinates to describe the configuration of the rigid body system. \cite{finzi2020simplifying} shows that choosing Cartesian coordinates and enforcing constraints explicitly simplifies the functional form of Hamiltonian and Lagrangian, which facilitates learning. Here we show a derivation based on D'Alembert's principle.

Assume the configuration of rigid bodies at time $t$ can be described by a set of Cartesian coordinates $\mathbf{x}(t) = (x_1(t), x_2(t), ..., x_D(t))$. From D'Alembert's principle, with a Lagrangian $L(\mathbf{x}, \dot{\mathbf{x}})$, Newton's second law can be expressed as 
\begin{equation}
\label{eq:D-cart}
    \delta \mathbf{x}^{T} \Big[\frac{\mathrm{d}}{\mathrm{d}t} \Big( \frac{\partial L}{\partial \dot{\mathbf{x}}} \Big) - \frac{\partial L}{\partial \mathbf{x}}\Big] = 0, 
\end{equation}
with the Lagrangian,
\begin{equation}
\label{eq:Lag-cart}
    L(\mathbf{x}, \dot{\mathbf{x}}) = \frac{1}{2} \dot{\mathbf{x}}^T \mathbf{M}\dot{\mathbf{x}} - V(\mathbf{x}),
\end{equation}
where the mass matrix $\mathbf{M} = \nabla_{\dot{\mathbf{x}}}\nabla_{\dot{\mathbf{x}}}^T L$ is a constant matrix that does not depend on the coordinates, which simplifies the functional form of Lagrangian. Please see \cite{finzi2020simplifying} for more details.
Since the $x_i$s are not independent, we need to incorporate the relationship between the $x_i$s - the constraints - to set up the equations of motion. In general, for a system of DOF $M$, there exists $D-M$ constraints. In this work, we assume all the constraints are holonomic, thus can be expressed as $K = D-M$ equality constraints $\Phi_k(\mathbf{x}) = 0$ for $k=1,2,..., K$. We collect functions $\Phi_k(\mathbf{x})$ into a column vector $\Phi(\mathbf{x})$ so the virtual displacements $\delta x_i$s are related by $K$ equations $(D_{\mathbf{x}}\Phi) \delta \mathbf{x} = \mathbf{0}$, where $D_{\mathbf{x}}\Phi$ denotes the Jacobian. We introduce $K$ Lagrangian multipliers $\lambda_k$s collected as a column vector $\lambda_L\in \mathbb{R}^K$ and we have  $\lambda_L^T (D_{\mathbf{x}}\Phi) \delta \mathbf{x} = 0$. Adding this scalar to Equation \eqref{eq:D-cart}, we get 
\begin{equation}
\label{eq:D-cart-con}
    \delta \mathbf{x}^{T} \Big[\frac{\mathrm{d}}{\mathrm{d}t} \Big( \frac{\partial L}{\partial \dot{\mathbf{x}}} \Big) - \frac{\partial L}{\partial \mathbf{x}} + (D_{\mathbf{x}}\Phi)^T  \lambda_L
    \Big] = 0.
\end{equation}
Even though the $\delta x_i$s are not independent, we can choose Lagrangian multipliers $\lambda_L$ such that the only solution to Equation \eqref{eq:D-cart-con} is one where the coefficients of all $\delta x_i$s are zeros. We symbolically expand the coefficients and get
\begin{equation}
\label{eq:EL-cart-con}
    \mathbf{M} \ddot{\mathbf{x}} + 
    \nabla_{\mathbf{x}} V  + (D_{\mathbf{x}}\Phi)^T  \lambda_L = \mathbf{0}.
\end{equation}
Note here we use the fact that $\nabla_{\mathbf{x}}\nabla_{\dot{\mathbf{x}}}^T L = \mathbf{0}$ and $\nabla_{\mathbf{x}} L = - \nabla_{\mathbf{x}} V$. The Lagrangian multipliers $\lambda_L$ can be solved as functions of $\mathbf{x}$ and $\dot{\mathbf{x}}$ by leveraging $\dot{\Phi} = (D_{\mathbf{x}}\Phi) \dot{\mathbf{x}} = \mathbf{0}$ and  $\ddot{\Phi} = (D_{\mathbf{x}}\Phi) \ddot{\mathbf{x}} +  (D_{\mathbf{x}}\dot{\Phi}) \dot{\mathbf{x}} = \mathbf{0}$ to eliminate $\ddot{\mathbf{x}}$. The solution is
\begin{equation}
\label{eq:lambda}
    \lambda_L = - [(D_{\mathbf{x}}\Phi) \mathbf{M}^{-1} (D_{\mathbf{x}}\Phi)^T ] ^{-1}
    [(D_{\mathbf{x}}\Phi) \mathbf{M}^{-1} \nabla_{\mathbf{x}} V - (D_{\mathbf{x}}\dot{\Phi}) \dot{\mathbf{x}}]
\end{equation}
Substitute the solved $\lambda_L$ into Equation \eqref{eq:EL-cart-con}, we get a set of $D$ second-order differential equations,
\begin{equation}
\label{eq:lag-constraint}
    \ddot{\mathbf{x}} = \mathbf{M}^{-1} (D_{\mathbf{x}}\Phi)^T [(D_{\mathbf{x}}\Phi) \mathbf{M}^{-1} (D_{\mathbf{x}}\Phi)^T ] ^{-1}
    [(D_{\mathbf{x}}\Phi) \mathbf{M}^{-1} \nabla_{\mathbf{x}} V - (D_{\mathbf{x}}\dot{\Phi}) \dot{\mathbf{x}}] - \mathbf{M}^{-1}  \nabla_{\mathbf{x}} V.
\end{equation}
%
%

%
%
\subsection{Hamiltonian dynamics with explicit constraints}
%
By applying Legendre transformation on the Lagrangian \eqref{eq:Lag-cart} we get the conjugate momentum $\mathbf{p}_{\mathbf{x}} = \nabla_{\dot{\mathbf{x}}} L = \mathbf{M} \dot{\mathbf{x}}$, and the Hamiltonian $H(\mathbf{x}, \mathbf{p}_\mathbf{x}) = \dot{\mathbf{x}}^T \mathbf{p}_\mathbf{x} - L(\mathbf{x}, \dot{\mathbf{x}})$, which has differential
\begin{equation}
\label{eq:ham-diff}
\delta H =  
\mathbf{p}_\mathbf{x}^T \delta \dot{\mathbf{x}} +  \dot{\mathbf{x}} ^T\delta \mathbf{p}_\mathbf{x}
- (\nabla_{\mathbf{x}} L)^T \delta\mathbf{x} - (\nabla_{\dot{\mathbf{x}}} L)^T \delta\dot{\mathbf{x}}
=
(\nabla_{\mathbf{x}} H)^T \delta\mathbf{x} +  (\nabla_{\mathbf{p}_{\mathbf{x}}} H)^T \delta\mathbf{p}_{\mathbf{x}}.
\end{equation}
The second equality holds because of the definition of $\delta H$. We can rewrite the equality as 
\begin{equation}
\label{eq:ham-diff2}
\begin{bmatrix}
\delta \mathbf{x}^T & \delta \mathbf{p}_{\mathbf{x}}^T
\end{bmatrix}
\begin{bmatrix}
\nabla_{\mathbf{x}} H + \nabla_{\mathbf{x}} L \\
\nabla_{\mathbf{p}_{\mathbf{x}}} H - \dot{\mathbf{x}}
\end{bmatrix}
=
\begin{bmatrix}
\delta \mathbf{x}^T & \delta \mathbf{p}_{\mathbf{x}}^T
\end{bmatrix}
\begin{bmatrix}
\nabla_{\mathbf{x}} H + \dot{\mathbf{p}}_\mathbf{x} \\
\nabla_{\mathbf{p}_{\mathbf{x}}} H - \dot{\mathbf{x}}
\end{bmatrix}
= 0.
\end{equation}
The first equality holds by substituting in D'Alembert's principle \eqref{eq:D-cart}. If all the coordinates are independent, we recover the Hamiltonian dynamics with implicit constraint \eqref{eq:ham}. The Cartesian coordinates are not independent so we need to explicitly add constraints with Lagrangian multipliers to derive equations of motion. Assume we have $K=D-M$ holonomic constraints collected into a vector equality $\Phi(\mathbf{x})=\mathbf{0}$, we can differentiate these constraints $\dot{\Phi}(\mathbf{x}) = (D_{\mathbf{x}}\Phi) \dot{\mathbf{x}} = (D_{\mathbf{x}}\Phi) \mathbf{M}^{-1} \mathbf{p}_{\mathbf{x}} = 0$ so that we form additional $K$ constraints that depend on both $\mathbf{x}$ and $\mathbf{p}_\mathbf{x}$. We denote $\mathbf{z} = (\mathbf{x}^T, \mathbf{p}_{\mathbf{x}}^T)^T$ and the $2K$ constraints as $\Psi(\mathbf{z}) = (\Phi^T, \dot{\Phi}^T)^T$. Then elements in the differential $\delta \mathbf{z}$ are related by $(D_{\mathbf{z}}\Psi)\delta \mathbf{z} = \mathbf{0}$. We introduce $2K$ Lagrangian multipliers $\lambda_H \in \mathbb{R}^{2K}$ and add the scalar $\lambda_H^T (D_{\mathbf{z}}\Psi) \delta \mathbf{z} = 0$ to \eqref{eq:ham-diff2}. We have
\begin{equation}
    \label{eq:ham-diff-con}
    \delta \mathbf{z}^T [\nabla_{\mathbf{z}} H + \mathbf{J} \dot{\mathbf{z}} + (D_{\mathbf{z}}\Psi)^T \lambda_H] = 0,
\end{equation}
where $\mathbf{J}$ is a symplectic matrix $\mathbf{J} = [\mathbf{0}, \mathbf{I}_D; -\mathbf{I}_D, \mathbf{0}]$ and $\mathbf{I}_D$ is the $D \times D$ identity matrix. 
Even though elements in $\delta \mathbf{z}$ are not independent, we can choose $\lambda_H$ such that the only solution to \eqref{eq:ham-diff-con} is one where the coefficients of all $\delta z_i$ are zeros. Thus we get $2D$ equations. To solve for $\lambda_H$ as a function of $\mathbf{z}$, we leverage $\dot{\Psi} =(D_{\mathbf{z}}\Psi)\dot{\mathbf{z}} = \mathbf{0}$ to eliminate $\dot{\mathbf{z}}$ and get 
\begin{equation}
    \label{eq:lambda_H}
    \lambda_H = - [(D_{\mathbf{z}}\Psi) \mathbf{J}(D_{\mathbf{z}}\Psi)^T ]^{-1}(D_{\mathbf{z}}\Psi) \mathbf{J} \nabla_{\mathbf{z}} H.
\end{equation}
Finally we get the equations of motion as $2D$ first order differential equations
\begin{equation}
    \label{eq:ham-con}
    \dot{\mathbf{z}} = \mathbf{J} \nabla_{\mathbf{z}} H - \mathbf{J}(D_{\mathbf{z}}\Psi)^T [(D_{\mathbf{z}}\Psi) \mathbf{J}(D_{\mathbf{z}}\Psi)^T ]^{-1}(D_{\mathbf{z}}\Psi) \mathbf{J} \nabla_{\mathbf{z}} H.
\end{equation}
%
%

%
%
\subsection{Learning of Dynamics}
%
All the dynamics derived above are first-order or second-order ordinary differential equations (ODE). The second-order ODEs, e.g., \eqref{eq:lag-constraint}, can be turned into a set of first-order ODEs by choosing state $\mathbf{s}$, e.g., $\mathbf{s} = (\mathbf{x}^T, \dot{\mathbf{x}}^T)^T$. If we know how to learn first-order ODEs from data, we will be able to learn Lagrangian or Hamiltonian dynamics of any form above. 

Neural ODE \cite{chen2018neural} is a framework for learning a general first-order ODE $\dot{\mathbf{s}} = \mathbf{f} (\mathbf{s})$ with an unknown function $\mathbf{f} (\mathbf{s})$. The idea is to parametrize $\mathbf{f} (\mathbf{s})$ by a neural network. With an initial condition $\mathbf{s}_0$, we can use an ODE solver to solve the initial value problem for future states,
\begin{equation}
    \hat{\mathbf{s}}_1, ..., \hat{\mathbf{s}}_T = \mathrm{ODESolve} (\mathbf{f}_{\theta}(\mathbf{s}), \mathbf{s}_0, 1, ..., T),
\end{equation}
where $\mathbf{f}_{\theta}(\mathbf{s})$ is the parametrization of $\mathbf{f}(\mathbf{s})$. We can then minimize the difference between the true states and the predicted states. As all the operations in the ODE Solver are differentiable, we can use back propagation to solve the optimization problem. To incorporate physics priors into this learning framework, we can parametrize unknown physical quantities in the derived dynamics above, e.g., mass, potential energy, the Lagrangian and the Hamiltonian. The difference between various recent models lies in what form of dynamics they use and what physical quantities they parametrize, which is explained in the next section. 
%
%
%

%
%
%
\section{Models Studied in this Survey}
%
Since the Lagrangian and Hamiltonian dynamics we consider here conserve energy, the neural network models that exploit Lagrangian and Hamiltonian dynamics automatically enforce energy conservation. In this section, we briefly summarize recent energy-conserving neural network models implemented in this benchmark study. As the original names of these models might not reflect the difference between them, we renamed a few models to emphasize the difference, while citing the original papers that introduced them.

\paragraph{HNN:}
\cite{greydanus2019hamiltonian} introduces Hamiltonian Neural Networks (HNN) to learn Hamiltonian dynamics \eqref{eq:ham} from data by parametrizing the Hamiltonian $H$ using a neural network. The original implementation assumes the data are in the form of $(\mathbf{q},\mathbf{p}, \dot{\mathbf{q}}, \dot{\mathbf{p}})_{0, ..., T}$. The learning is performed by feeding data $(\mathbf{q},\mathbf{p})$ into the Hamiltonian neural network to get the scalar $H(\mathbf{q},\mathbf{p})$, getting the right hand side (RHS) of \eqref{eq:ham} by automatic differentiation, and minimizing the violation of the dynamics \eqref{eq:ham} with data $(\dot{\mathbf{q}}, \dot{\mathbf{p}})$. In this benchmark, we assume we can only access position and velocity data. Thus, in our implementation, we additionally parametrize the mass matrix $\mathbf{M}(\mathbf{q})$ using a neural network and leverage Neural ODE to learn the dynamics. 

\paragraph{HNN-structure:}
For rigid body systems, instead of parametrizing the Hamiltonian, we can leverage the structure of the Hamiltonian \eqref{eqn:Ham} and parametrize the mass matrix $\mathbf{M}(\mathbf{q})$ and potential energy $V(\mathbf{q})$ using neural networks. This allows us to learn Hamiltonian dynamics in the form of \eqref{eq:1st-order-ham-s}. This learning scheme is introduced as Symplectic ODE-Net (SymODEN) \citep{Zhong2020Symplectic}, which can jointly learn an additional control term. Dissipative SymODEN \citep{zhong2020dissipative} jointly learns an additional dissipation term. Symplectic Recurrent Neural Network (SRNN) \citep{Chen2020Symplectic} also exploits the structure of the Hamiltonian by assuming a separable Hamiltonian.

\paragraph{CHNN:} 
By explicitly enforcing constraints,  Hamiltonian dynamics in the form of \eqref{eq:ham-con} can be learned from Cartesian coordinate data. Since the mass matrix under Cartesian coordinates is constant and sparse, elements in the mass matrix can be set as learnable parameters. Moreover, for a system of multiple rigid bodies, the mass matrix is block diagonal, which further reduces the number of learnable parameters \citep{finzi2020simplifying}. As the inverse of mass matrix appears all the forms of Hamiltonian dynamics, the block diagonality and sparsity of the constant mass matrix under Cartesian coordinates facilitates the computation of matrix inverse. This learning scheme is introduced as Constrained HNN (CHNN) in \cite{finzi2020simplifying}. 

\paragraph{LNN:}
\cite{cranmer2020lagrangian} introduces Lagrangian Neural Networks (LNN) to learn Lagrangian dynamics \eqref{eq:2nd-order-lag} from position and velocity data by parametrizing the Lagrangian $L$ using a neural network. The benefit of this learning scheme is that it can learn any arbitrary Lagrangian, which does not need to be quadratic in velocity as in rigid body systems. 

\paragraph{LNN-structure:}
For rigid body systems, the structure of the Lagrangian \eqref{eq:Lag} can be leveraged and the mass matrix $\mathbf{M}(\mathbf{q})$ and potential energy $V(\mathbf{q})$ can be parametrized using neural networks to learn Lagrangian dynamics in the form of \eqref{eq:2nd-order-lag-s}. This learning scheme is first introduced as Deep Lagrangian Network (DeLaN) in \cite{lutter2018deep} which jointly learn an additional control term for the purpose of controlling robotic systems. DeLaN is also used in \cite{lutter2019deep} to learn and control under-actuated systems. DeLaN assumes that position $\mathbf{q}$, velocity $\dot{\mathbf{q}}$ and acceleration $\ddot{\mathbf{q}}$ data can be accessed from sensor and the violation of \eqref{eq:2nd-order-lag-s} is minimized. 
For the purpose of bench-marking, we leverage Neural ODE to learn \eqref{eq:2nd-order-lag-s} without acceleration data. 

\paragraph{CLNN:}
Learning Lagrangian dynamics with explicit constraints in the form of \eqref{eq:lag-constraint} from Cartesian coordinate data is introduced as Constrained LNN (CLNN) in \cite{finzi2020simplifying}. Similar to CHNN, the mass matrix properties of block diagonality and sparsity in each block are exploited to facilitate learning. 
\paragraph{HNN-angle, HNN-structure-angle, LNN-angle, LNN-structure-angle:} 
When enforcing constraints implicitly, we will often encounter angular coordinates. If we treat each angular coordinate as a variable in $\mathbb{R}^1$, the learning algorithm would not be able to infer, for example, $0$ and $2\pi$ represents the same angle. In fact, each angular coordinate lies on the manifold $\mathbb{S}^1$ (circle) instead of manifold $\mathbb{R}^1$ (line).  This problem can be solved by embedding an angular coordinate $q$ as $(\cos q, \sin q) \in \mathbb{S}^1$. Thus, for the four models that enforce constraints implicitly (HNN, HNN-strucure, LNN, LNN-structure), we also implement their angle-aware version (names with ``angle" suffix) by embedding each angular coordinate into $\mathbb{S}^1$ before feeding it into the neural networks. This angle-aware design is also implemented and studied in SymODEN \citep{Zhong2020Symplectic}.
\paragraph{}For models above that parametrize the mass matrix $\mathbf{M}({\mathbf{q}})$ using a neural network, we leverage the fact that for a real physical system, the mass matrix is positive definite and can be expressed as $\mathbf{M}({\mathbf{q}}) = \mathbf{L}({\mathbf{q}}) \mathbf{L}({\mathbf{q}})^T$ (Cholesky decomposition), where $\mathbf{L}({\mathbf{q}})$ is a lower triangular matrix. Thus, instead of parametrizing the mass matrix or the inverse of mass matrix, we parametrize $\mathbf{L}({\mathbf{q}})$, as done in \cite{lutter2018deep} and \cite{Zhong2020Symplectic}.

\section{Experiments}

\subsection{Experimental Setup}
\begin{wrapfigure}[5]{r}{0.48\textwidth}
    \vspace{-8em}
    \centering
    \includegraphics[width=0.22\textwidth]{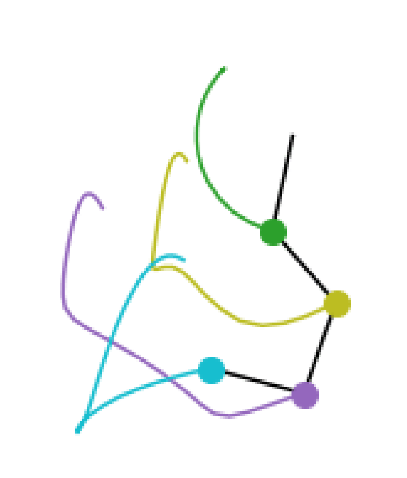}
    \includegraphics[width=0.21\textwidth]{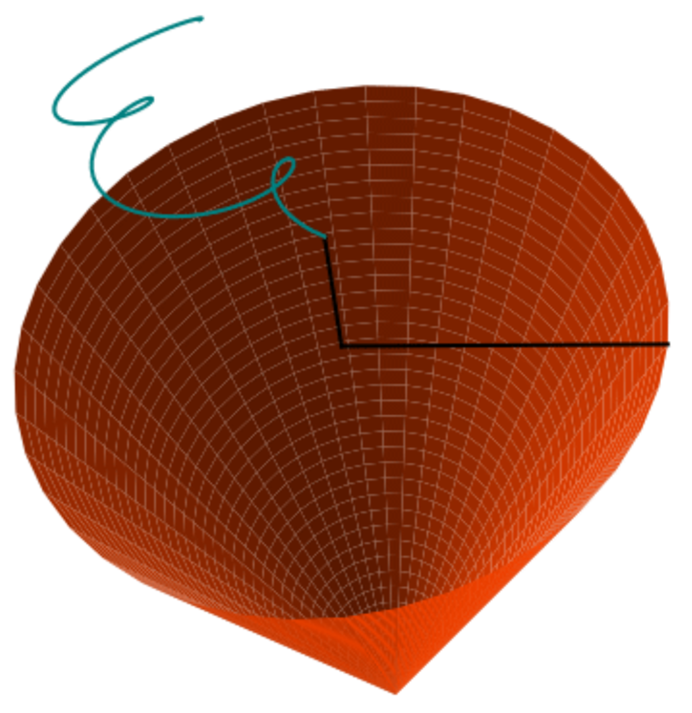}
    \vspace{-1em}
    \caption{\textbf{Left:} 4-pendulum; \textbf{Right:} gyroscope}
    \label{fig:gyro}
    \vspace{-2em}
\end{wrapfigure}
Figure \ref{fig:gyro} shows an N-pendulum system and a gyroscope. The N-pendulum is a set of 2D systems. The complexity of the dynamics increases as the number of pendulums increases. Except for the 1-pendulum, the systems are chaotic system which means accurate prediction is not possible when $N \geq 2$ \citep{kwok2020modeling}.  The gyroscope is a 3D system that exhibits complex dynamics including precession and nutation \citep{butikov2006precession}. In this work, we simulate the 1-, 2- and 4-pendulum systems and the gyroscope to evaluate models.

\paragraph{Data Generation and Training} For each of the 4 physical systems, we randomly generate 800 initial conditions of states and roll out simualtion trajectories for 100 time steps for training. We randomly generate another set of 100 initial conditions of 100 time steps for testing. Random Gaussian noises of standard deviation 0.01 are added independently to every coordinate and velocity. For the N-Pendulum systems, we use data in 2D Cartesian coordinates to train CHNN and CLNN, and data in planar angular coordinates to train the rest of the models. For the gyroscope, we use data in 3D Cartesian coordinates of 4 chosen points in the gyroscope as done in \citep{finzi2020simplifying} to train CHNN and CLNN, and data in Euler angles to train the rest of the models. We generate trajectories using RK4 integrator instead of symplectic integrators as done in \cite{zhu2020deep}, \cite{sanchez2019hamiltonian}, \cite{Chen2020Symplectic}, \cite{jin2020sympnets}, \cite{tong2020symplectic} because using RK4 integrator allows us to incorporate control and dissipation into the framework, whereas symplectic integrators does not. For the N-pendulum systems and the gyroscope system, the integration interval are 0.03s and 0.02s, respectively. As function approximators, we use fully-connected feedforward network with 3 hidden layers with 256 neurons/layers and Softplus activation. For learnable mass/moments in CLNN/CHNN, we implement them as learnable parameters passed through an exponential function, since the mass/moments are always positive for real physical systems. All models are trained with \emph{AdamW} optimizer \cite{loshchilov2018decoupled} for 1000 epochs and a learning rate of 0.001.

\paragraph{Metrics} We use $L(\mathbf{s}_{(0,...T)}, \hat{\mathbf{s}}_{(0,...,T)}) = \sum_{t = 0}^T ||\hat{\mathbf{s}}_t - \mathbf{s}_t ||_1$ as the loss function for training, since it has been shown that the robustness of $L_1$ loss to outliers benefits learning complex dynamics \citep{finzi2020simplifying}. While training, we set $T=4$ and select random snapshots of length 5 from each long trajectory. We evaluate the performance of each model by the \emph{relative error} at each time step  $||\hat{\mathbf{s}}_t - \mathbf{s}_t ||_2 / || \mathbf{s}_t ||_2$ and the \emph{trajectory relative error} $ \sum_{t=0}^T||\hat{\mathbf{s}}_t - \mathbf{s}_t ||_2 / || \mathbf{s}_t ||_2$, which sums up relative errors along a trajectory. We also analyze each model's ability in conserving true energy with the \emph{absolute error} in true energy $||\hat{E}_t - E_t ||_2$.  Since the absolute value of energy is not meaningful (potential energy is relative to an arbitrarily chosen zero position), we do not choose the relative error in true energy as the metric, which scales the absolute error by the absolute value of energy.

\subsection{Results}
\begin{figure}%
    \centering
    \includegraphics[width=0.49\linewidth]{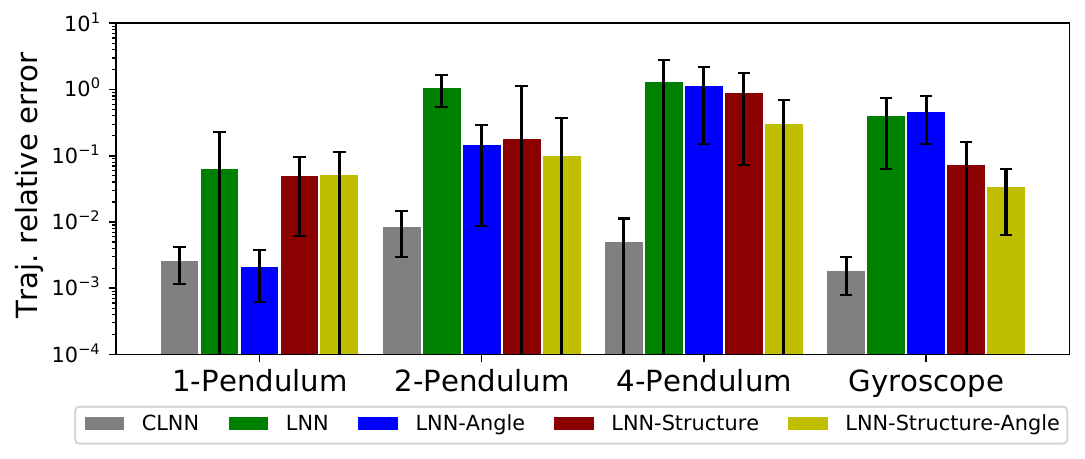}%
    \includegraphics[width=0.49\linewidth]{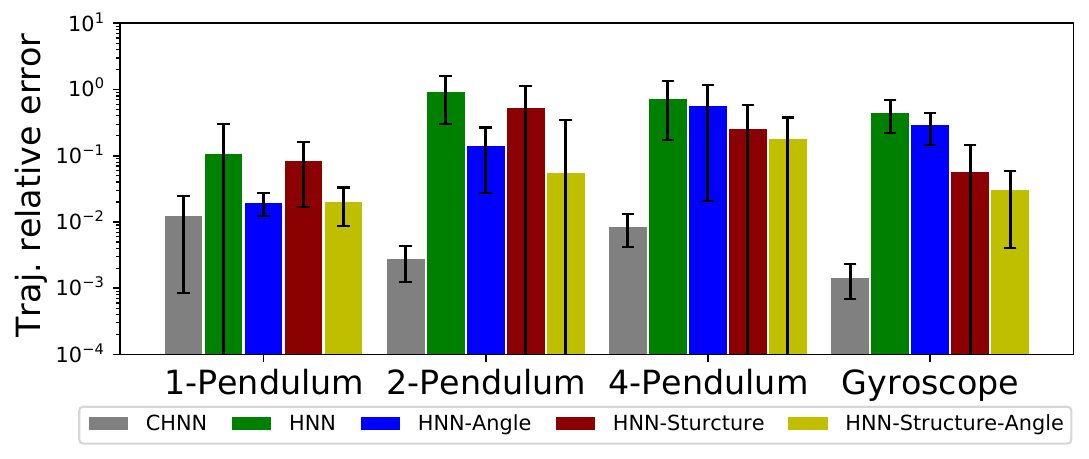}%
    \vspace{-1.0em}
    \caption{The trajectory relative error (log scale) in the states ($T=4$). Each error is averaged over trajectories with 100 test initial conditions. The absolute error bars show the standard deviation in \textit{log} scale. \textbf{Left}: Lagrangian models; \textbf{Right}: Hamiltonian models.}
    \label{fig:test_err}
    \vspace{-1em}
\end{figure}
\begin{figure}
    \centering
    \includegraphics[width=0.98\linewidth]{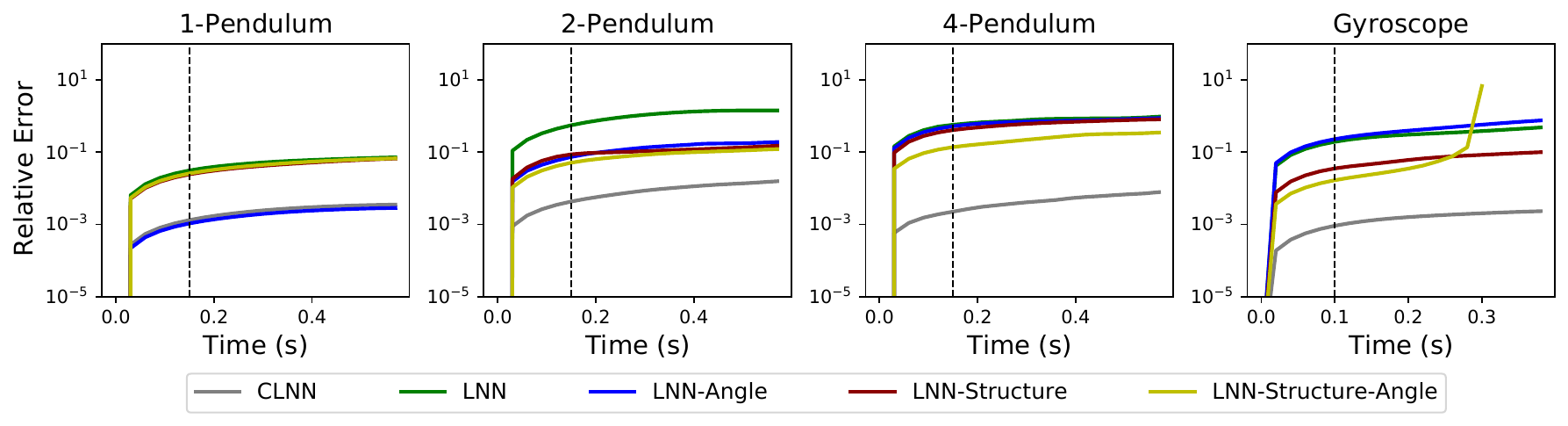}
    \includegraphics[width=0.98\linewidth]{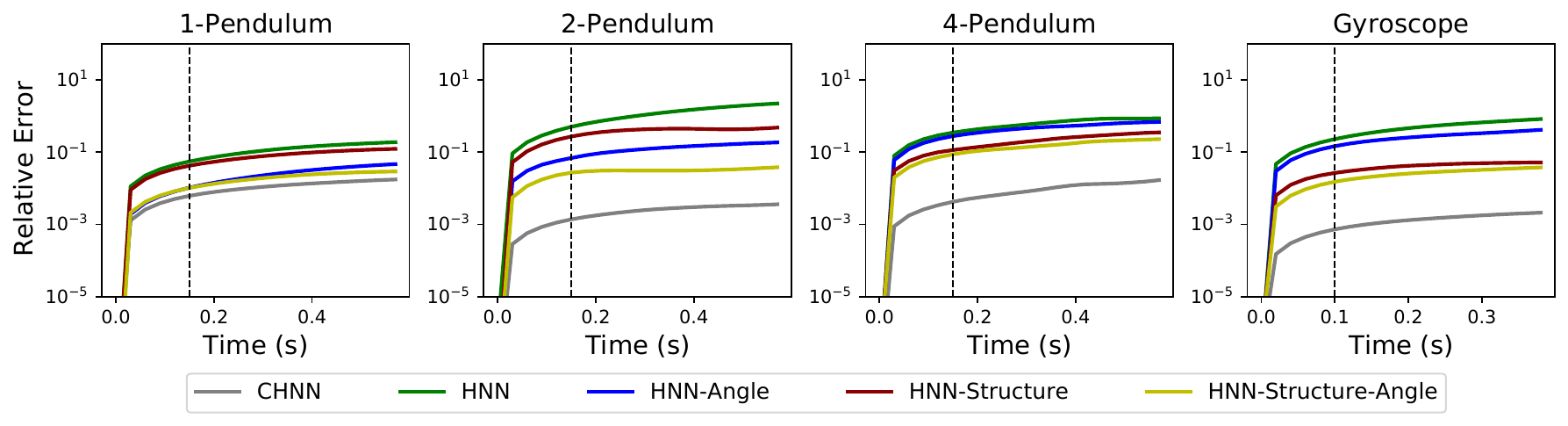}
    \vspace{-1.5em}
    \caption{The relative error in the states (log scale). Each curve is averaged over the prediction trajectories of 100 test initial conditions. The vertical dashed lines shows the trajectory length in training. \textbf{Top}: Lagrangian models; \textbf{bottom}: Hamiltonian models. }
    \label{fig:pred_err}
    \vspace{-1.5em}
\end{figure}
\paragraph{Test error} Figure \ref{fig:test_err} shows the trajectory relative error of 10 models on test trajectory data. The models with explicit constraints (CLNN and CHNN) consistently outperforms all the other models across all systems. The gap is the largest in the 3D system (Gyroscope) since with explicit constraints, the dynamics under Cartesian coordinates do not suffer from gimbal lock, which occurs in the dynamics under Euler angles (implicit constraints). For the models with implicit constraints, we can observe the benefit of leveraging a structured Lagrangian/Hamiltonian and angle-aware design. Comparing the angle-aware models with their non-angle-aware variants (e.g., LNN-Angle and LNN), we conclude that the angle-aware design in general benefits learning. This conclusion is consistent with our intuition since respecting the geometry of coordinates lets models identify $q$ and $q\pm 2\pi$, which helps models learn dynamics. Comparing models that leverage structured Lagrangian and those that do not (e.g., LNN-Structure and LNN), we observe that the structure benefits the learning of 2 and 4-Pendulum systems while does not help the learning of 1-Pendulum and Gyroscope. The 1-Pendulum has simple dynamics, so it can be well learned without the structure. As the number of pendulums increases, the structure in Lagrangian/Hamiltonian helps recover dynamics that explain the data. For 3D Gyroscope, it is unclear why the structure does not help. We leave it as an open question for future work.
\paragraph{Long-term prediction.} We evaluate 10 models on trajectories longer than those in training ($T=4$). Figure \ref{fig:pred_err} shows the average relative error over 100 test trajectories of length $T=20$. The vertical dashed line shows the length of training trajectories. 
The 2-Pendulum and 4-Pendulum are chaotic systems so accurate long-term prediction is not possible. In general, CHNN and CLNN performs the best across all systems considered. We have similar observations on the role of angle-aware design and the structured Lagrangian/Hamiltonian.
\begin{figure}
    \centering
    \includegraphics[width=0.98\linewidth]{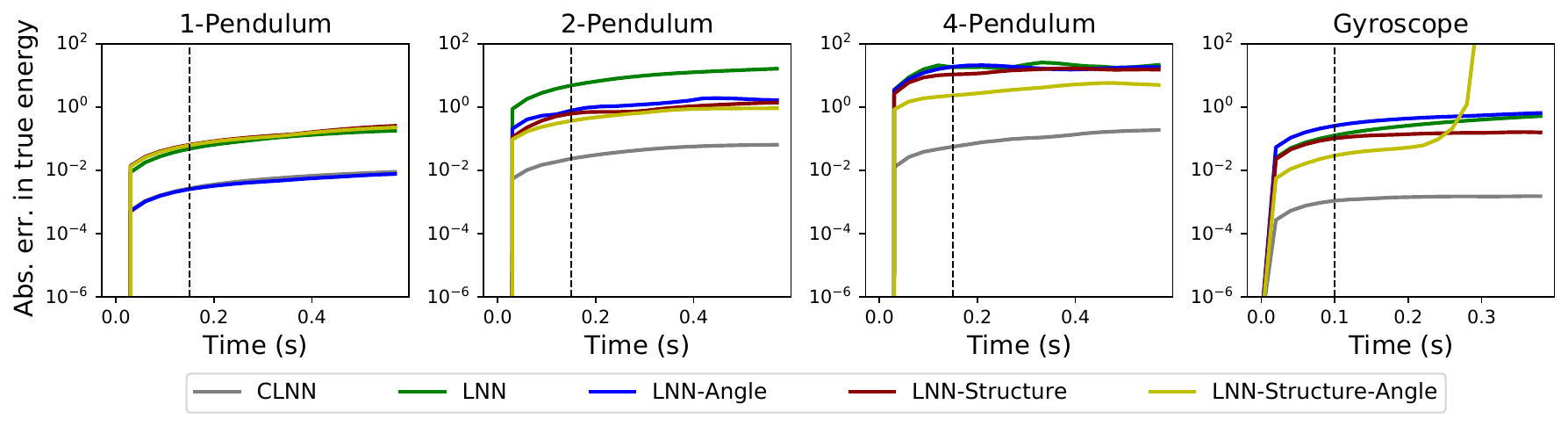}
    \includegraphics[width=0.98\linewidth]{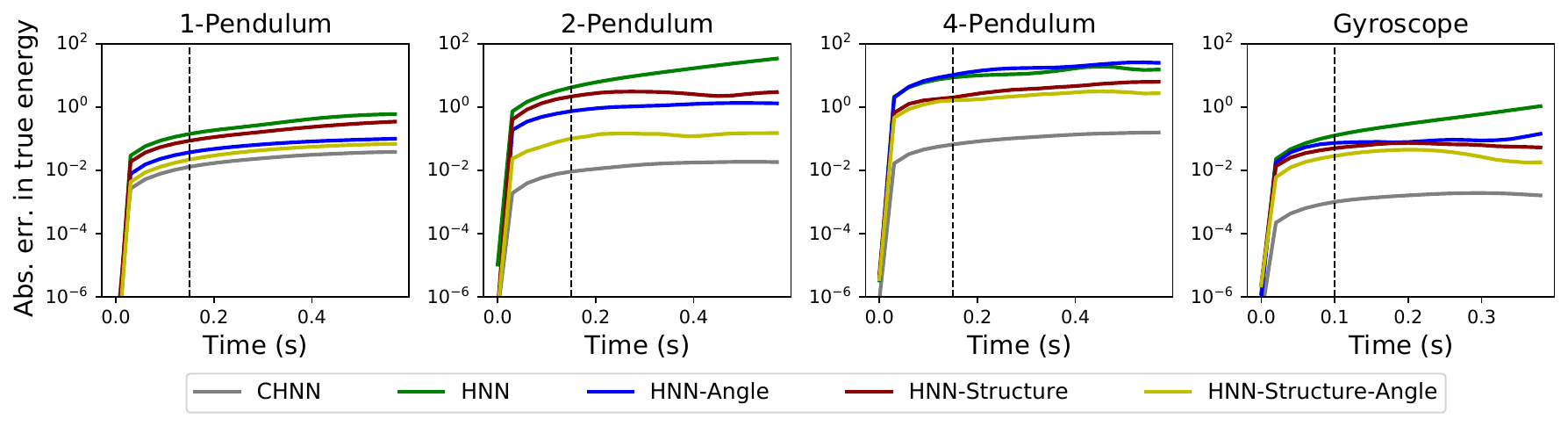}
    \vspace{-1.5em}
    \caption{Absolute error in true energy (log scale). Each curve is averaged over the prediction trajectories of 100 test initial conditions. The vertical dash lines shows the trajectory length in training. \textbf{Top}: Lagrangian models; \textbf{bottom}: Hamiltonian models.}
    \label{fig:energy_err}
    \vspace{-1.5em}
\end{figure}
\paragraph{Energy conservation} All 10 models naturally incorporate the prior of energy conservation, which means the learned energy is conserved. However, the learned energy might not accurately represent the true energy. Figure \ref{fig:energy_err} shows how the true energy calculated from prediction trajectories differ from the constant true energy along ground truth trajectories. We observe that although each model conserves its own learned energy, they all drift away from the constant true energy as time goes by. CLNN and CHNN has relative low drifts as compared to other models, which explains why they perform better in prediction. 
\begin{figure}%
    \centering
    \includegraphics[width=0.49\linewidth]{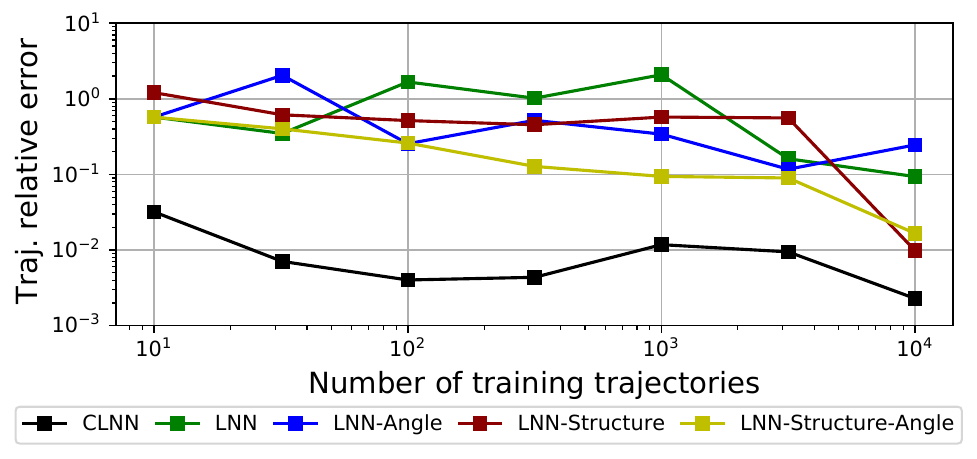}%
    \includegraphics[width=0.49\linewidth]{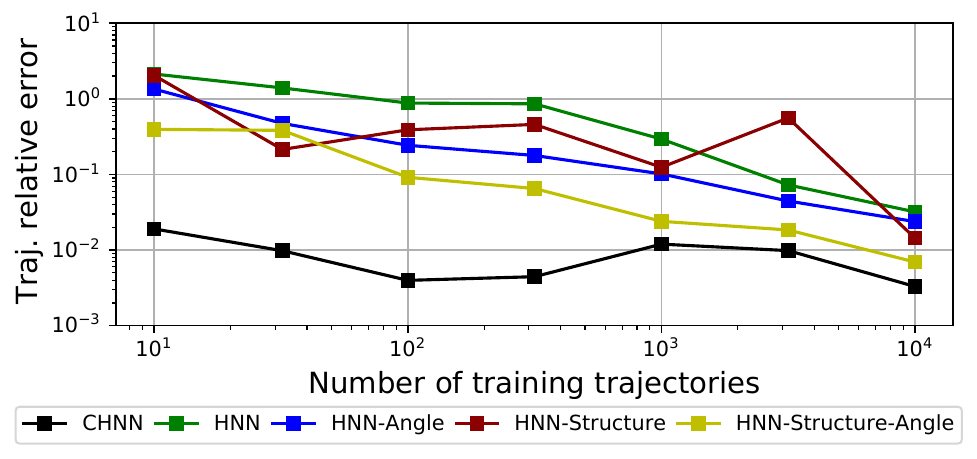}%
    \vspace{-1em}
    \caption{Trajectory relative error in the states (log scale) of 2-Pendulum with various sizes of training trajectories. Each curve is averaged over 100 test trajectories ($T=4$). \textbf{Left}: Lagrangian models; \textbf{Right}: Hamiltonian models.}
    \vspace{-1em}
    \label{fig:data_eff}%
\end{figure}
\paragraph{Data Efficiency} To compare the data efficiency of 10 models, we change the size of training dataset on the 2-Pendulum system from as low as 10 trajectories to as high as 10,000 trajectories, as shown in Figure \ref{fig:data_eff}. We observe that CLNN and CHNN outperforms all the other models. We also find that prior knowledge (structure/angle-aware) benefits data efficiency.
\section{Conclusion and Discussion}
In this work, we benchmark recent energy-conserving neural network models that rely on Lagrangian/Hamiltonian dynamics. CLNN and CHNN, which enforce constraints explicitly, have a significant advantage in prediction over models that enforce constraints implicitly as observed in \cite{finzi2020simplifying}, because the Lagrangian and Hamiltonian assume a much simpler form in Cartesian coordinates. This is due to the fact that the mass matrix under Cartesian coordinates is coordinate-independent, whereas the mass matrix depends on the coordinates in a complicated way when we use angular coordinates to describe the system. Moreover, for those models that need to learn a coordinate-dependent mass matrix (all Hamiltonian models except CHNN, and LNN-Structure, LNN-Structure-Angle), the initialized parametrization of the mass matrix tend to be ill-conditioned, which makes training unstable. In practice, a small positive constant is added to the diagonal elements of the mass matrix parametrization to alleviate this problem; however, this caveat complicates the training of these implicit models. For LNN and LNN-Angle, which do not learn the mass matrix, it is implicitly derived from the Hessian of the Lagrangian; since computing Hessian involves two gradient operations, these two models incur an additional computational cost and result in slower training. The advantage of models that enforce constraints implicitly lies in their easy implementation, whereas implementing CHNN and CLNN requires manually adding constraints for the systems of interest. 

We also emphasize the similarity and difference between the analytical forms that the models learn and the intuition behind them. We conclude that angle-aware design in general benefits learning and imposing structures on the Lagrangian and the Hamiltonian benefits learning for 2D systems. In fact, CLNN and CHNN also leverage the structure in Lagrangian and Hamiltonian to simplify the dynamics. As the structure can be leveraged to design energy-based controllers \citep{Zhong2020Symplectic}, we believe it is promising to design controllers based on CLNN and CHNN. 

\section*{Acknowledgement}
The authors would like to thank Marc Finzi, Ke Alexander Wang, and Andrew Gordon Wilson for their helpful comments on the manuscript. The authors also find their codebase released along with \cite{finzi2020simplifying} helpful when conducting the experiments in this survey.

\bibliography{ref}

\end{document}